\documentclass{article}

\usepackage{microtype}
\usepackage{graphicx}
\usepackage{subcaption}
\usepackage{booktabs} 

\usepackage{multicol}
\usepackage{multirow}
\usepackage{makecell}

\usepackage{hyperref}
\usepackage{dsfont}

\usepackage[preprint]{icml2026}

\usepackage{amsmath}
\usepackage{amssymb}
\usepackage{mathtools}
\usepackage{amsthm}
\usepackage{gensymb}

\usepackage[capitalize,noabbrev]{cleveref}

\theoremstyle{plain}

\theoremstyle{definition}

\theoremstyle{remark}

\usepackage[textsize=tiny]{todonotes}

\icmltitlerunning{WorldComp2D}

\begin{document}

\twocolumn[
  \icmltitle{WorldComp2D: Spatio-semantic Representations of Object Identity and Location from Local Views}



  \icmlsetsymbol{equal}{*}

  \begin{icmlauthorlist}
    \icmlauthor{SeongMin Jin}{a}
    \icmlauthor{Doo Seok Jeong}{a,b}
  \end{icmlauthorlist}

  \icmlaffiliation{a}{Division of Materials Science and Engineering}
  \icmlaffiliation{b}{Department of Semiconductor Engineering, Hanyang University, Republic of Korea}
  \icmlcorrespondingauthor{Doo Seok Jeong}{dooseokj@hanyang.ac.kr}

  \icmlkeywords{Machine Learning, ICML}
  \vskip 0.3in
]

\printAffiliationsAndNotice{}  

\begin{abstract}
Learning latent representations that capture both semantic and spatial information is central to efficient spatio-semantic reasoning. However, many existing approaches rely on implicit latent structures combined with dense feature maps or task-specific heads, limiting computational efficiency and flexibility.
We propose WorldComp2D, a novel lightweight representation learning framework that explicitly structures latent space geometry according to object identity and spatial proximity using multiscale \textit{local} receptive fields. This framework consists of (i) a proximity-dependent encoder that maps a given observation into a spatio-semantic latent space and (ii) a localizer that infers the coordinates of objects in the input from the resulting spatio-semantic representation.
Using facial landmark localization as a proof-of-concept, we show that, compared to SoTA lightweight models, WorldComp2D reduces the numbers of parameters and FLOPs by up to $4.0\times$ and $2.2\times$, respectively, while maintaining real-time performance on CPU. These results demonstrate that explicitly structured latent spaces provide an efficient and general foundation for spatio-semantic reasoning. This framework is open-sourced at https://github.com/JinSeongmin/WorldComp2D.
\end{abstract}

\section{Introduction}
\begin{figure}[tb]
  \begin{center}
    \centerline{\includegraphics[width=\columnwidth]{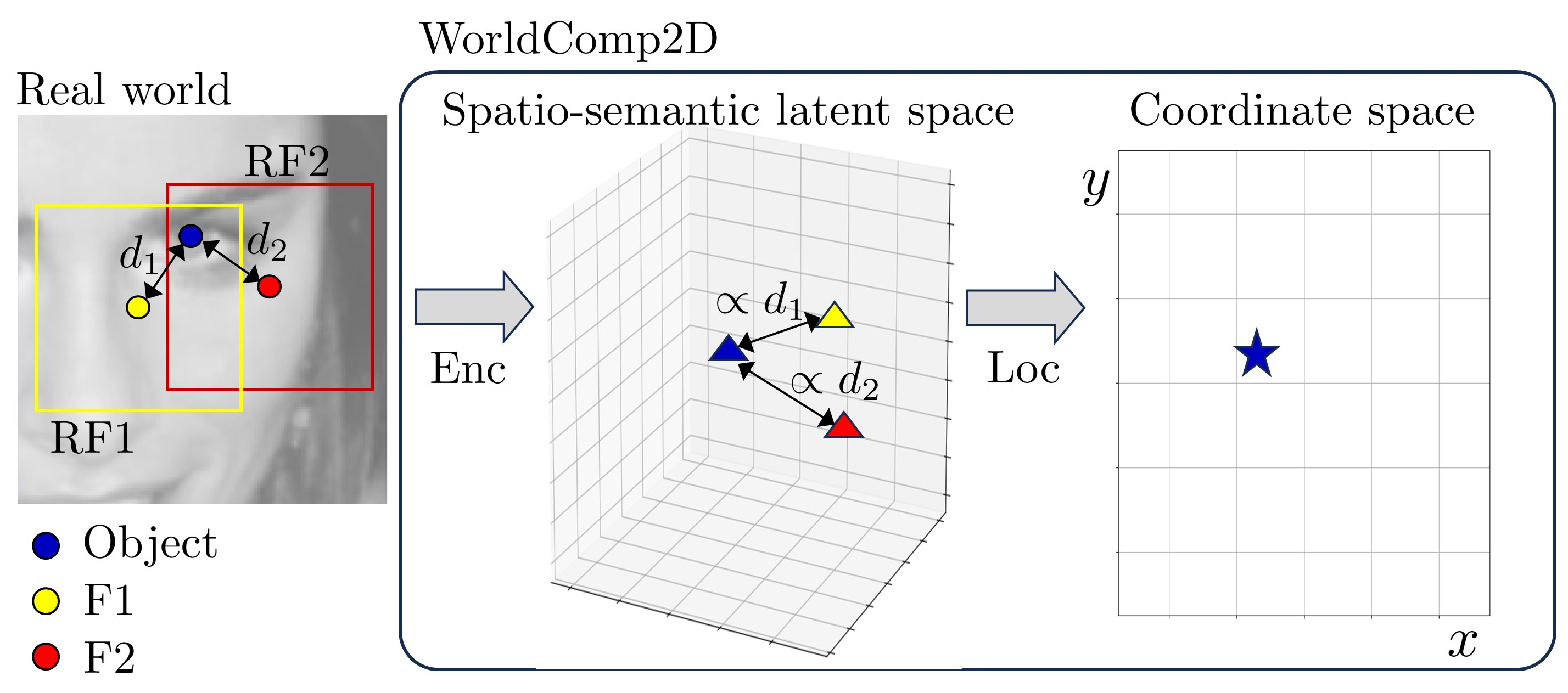}}
    \caption{\label{fig:framework} Overview of WorldComp2D. Observations made by an agent are encoded into a spatio-semantic latent space in which object identity is preserved and latent distances reflect real-world spatial proximity. Object locations are then inferred from these representations via the localizer. RF1 and RF2 denote two receptive fields centered at fixation points F1 and F2, respectively.}
  \end{center}
  \vskip -0.1in
\end{figure}

Embodied artificial intelligence (AI) agents may need to rely on internal latent representations to encode high-dimensional and complex real-world visual input into compact, structured forms that summarize the underlying factors of variation~\cite{ha2018world}. Unlike conventional vision systems, embodied agents operate under fundamentally different conditions: instead of having access to a complete global view of the environment, they perceive the world through local observations dependent on viewpoints~\cite{papoudakis2021partial}. To operate in real-time, such agents must rapidly construct internal representations that support instance-level understanding from these local views while operating under strict constraints on computation and power consumption.

Despite significant progress in representation learning, most existing approaches to spatio-semantic inference, which require jointly reasoning about semantic identity and spatial location such as object localization, are designed around global image access. These methods typically process full-images through pixel-wise computation~\cite{ren2016faster, lin2017focal}, specialized architectures~\cite{newell2016stacked}, or heatmap-based regression heads~\cite{wang2020hrnet, huang2021adnet}. Although effective in standard computer vision benchmarks, they fundamentally assume a complete global view of the environment, which conflicts with embodied AI settings. Moreover, full-image processing incurs substantial computational cost and latency, which limits their suitability for real-time deployment constraints.

In this work, we propose WorldComp2D, a novel lightweight representation learning framework for embodied AI. Our key idea is to directly encode spatio-semantic structure into the latent space. As illustrated in Fig.~\ref{fig:framework}, observations made by an agent are encoded into a latent space where object identity is preserved and latent distances reflect real-world spatial proximity. Then, a localizer predicts object locations directly from these representations. This design enables efficient spatio-semantic inference from a small set of local observations, consistent with the perceptual and computational constraints of embodied agents. 

We evaluate WorldComp2D using facial landmark localization as a proof-of-concept. While achieving slightly lower accuracy than SoTA regression methods, our framework enables robust inference from limited visual observations with substantially reduced computational complexity. WorldComp2D requires only $2.4$M parameters and $<550$ MFLOPs, achieving $>78$ frames per second (FPS) on CPU, whereas prior approaches typically rely on significantly larger models with $9.7-67$M parameters and $1.2-26.8$ GFLOPs.

Our contributions are as follows:
\begin{itemize}   
    \item We propose WorldComp2D, a lightweight fixation-centered framework that learns spatio-semantic representations from local views and localizes objects using an encoder and a localizer, with an optional refinement module.
    \item We design a learning objective that preserves object identity while encouraging latent-space distances to reflect real-world distances to the fixation point through proximity-weighted contrastive learning.
    \item We demonstrate that WorldComp2D achieves competitive localization accuracy with real-time efficiency and flexible accuracy–speed trade-offs, supported by extensive ablation and robustness studies.
\end{itemize}

\section{Related Work}
\subsection{Representation Learning for Embodied AI}
In embodied AI, latent representations serve as internal states that summarize sensory observations under partial observability, as agents perceive the environment through local, viewpoint-dependent inputs. In this context, previous work has explored world model-based approaches that map high-dimensional observations to compact latent states, where a learned transition function predicts their evolution over time for imagination, planning, and control~\cite{ha2018world, hafner2019learning}. These approaches enable agents to predict how the environment changes over time under their actions. Although effective for sequential decision making, the latent space is primarily shaped by temporal prediction objectives and is not explicitly structured to preserve fine-grained spatial or instance-level information.

In contrast, WorldComp2D is designed to directly support embodied perception by explicitly structuring latent space geometry. Rather than focusing on temporal dynamics, our framework encodes the semantic identity of objects and fixation-centered spatial proximity as intrinsic properties of the representation. By embedding local view-dependent semantic and spatial information directly into latent space, WorldComp2D enables efficient spatio-semantic inference from a small set of local observations, providing an internal representation that is naturally aligned with embodied perception.

\subsection{Contrastive Learning}
Contrastive learning has emerged as a powerful paradigm for learning discriminative representations by encouraging similarity between related samples while pushing apart unrelated ones. In a typical setting, two augmented views are generated for each sample, resulting in $2N$ views for a mini-batch of $N$ samples. Based on these augmented views, contrastive learning methods define binary positive and negative pairs according to instance identity, data augmentations, or semantic labels~\cite{chen2020contrastive, tian2020contrastive, khosla2020supcon}.

Our approach departs from conventional binary contrastive formulations by modeling pairwise relationships with continuous affinity values. Rather than assigning equal importance to all positive pairs, we incorporate spatial proximity as a continuous signal that modulates the strength of pairwise interactions. This design enables a more precise and fine-grained notion of distance in the latent space, allowing representations to capture fixation-centered proximity beyond categorical similarity.

\subsection{Object Localization}
Object localization has been widely studied for identifying and localizing objects or keypoints in images. Early approaches depend on full-image processing and pixel-wise prediction schemes, such as anchor-based detectors~\cite{ren2016faster, lin2017focal}. While effective, these approaches often involve high computational cost and strong task-specific architectural assumptions.

Facial landmark localization can be viewed as a structured instance of object localization, in which semantically meaningful keypoints are estimated under fixed spatial constraints~\cite{zhang2014facial, bulat2017far}. Most existing facial landmark localization methods employ heatmap-based regression as the primary mechanism for spatial inference~\cite{wang2019awing, wang2020hrnet}, typically requiring high-resolution feature maps and multi-stage refinement, which further increases computational cost.

WorldComp2D enables object localization without full-image processing. Rather than performing pixel-wise computations, our framework encodes a small set of local observations into latent representations and infers object locations by aggregating the resulting information directly in latent space. This design allows localization to be achieved efficiently from limited local views, without relying on exhaustive image-level processing. Conventional heatmap-based regression is optionally employed as a lightweight refinement mechanism, providing improved accuracy.

\section{Methodology}
\begin{figure*}[tb]
  \begin{center}
    \centerline{\includegraphics[width=1.7\columnwidth]{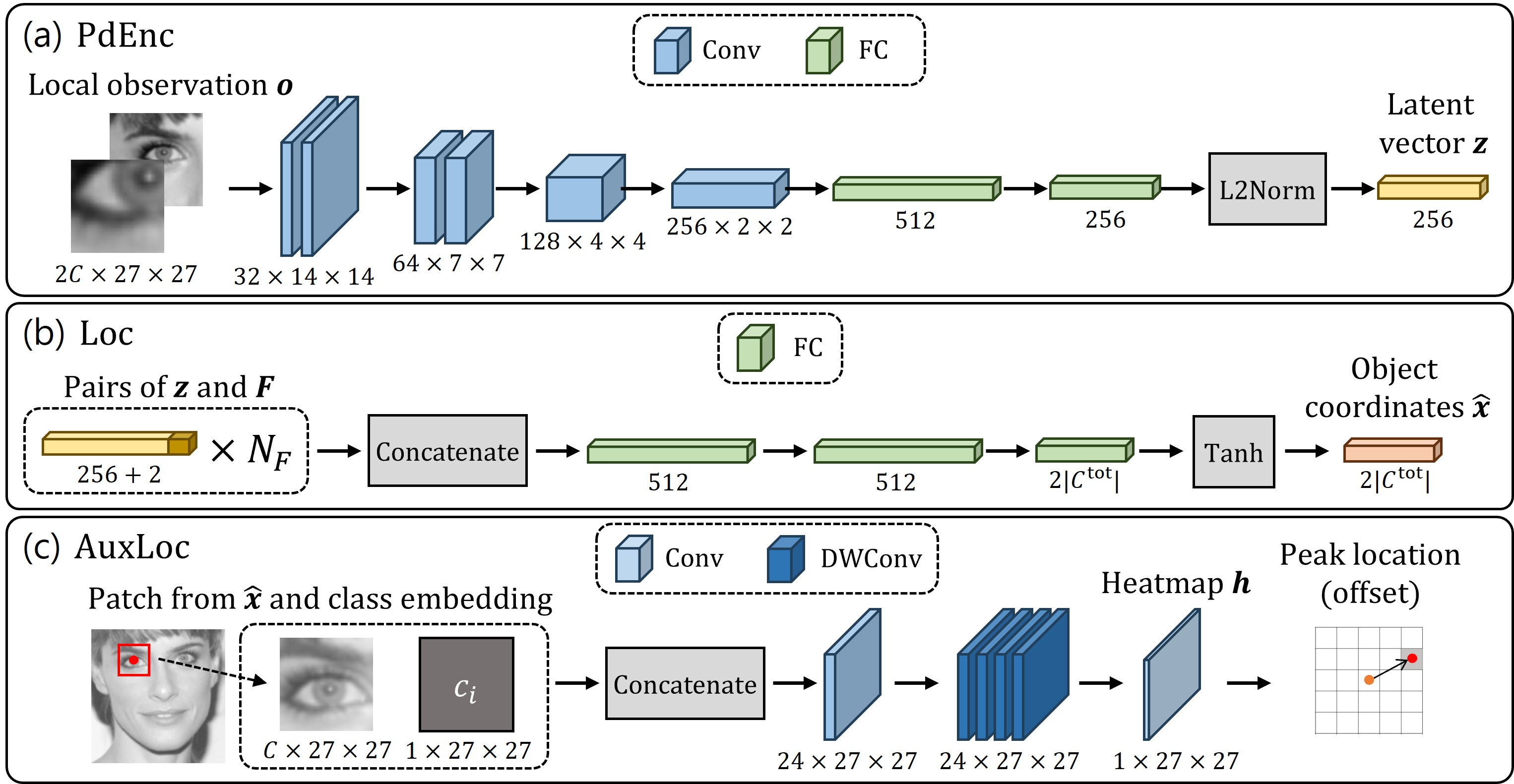}}
    \caption{\label{fig:architectures} Networks in WorldComp2D. (\textbf{a}) Proximity-dependent encoder (PdEnc), which maps fixation-centered observations to a normalized latent vector. (\textbf{b}) Localizer (Loc), which aggregates paired latent vectors and fixation coordinates to predict object locations. (\textbf{c}) Auxiliary localizer (AuxLoc), an optional refinement module that estimates a heatmap from a local patch and class-conditioned embedding.}
  \end{center}
  \vskip -0.1in
\end{figure*}

WorldComp2D comprises a proximity-dependent encoder ($\textrm{PdEnc}$) and a localizer ($\textrm{Loc}$), which consider objects belonging to a set of given classes. WorldComp2D also includes an optional auxiliary localizer ($\textrm{AuxLoc}$) to improve object localization accuracy. These three models are illustrated in Fig.~\ref{fig:architectures}.

\subsection{Proximity-dependent Encoder}
\begin{figure}[tb]
  \begin{center}
    \centerline{\includegraphics[width=\columnwidth]{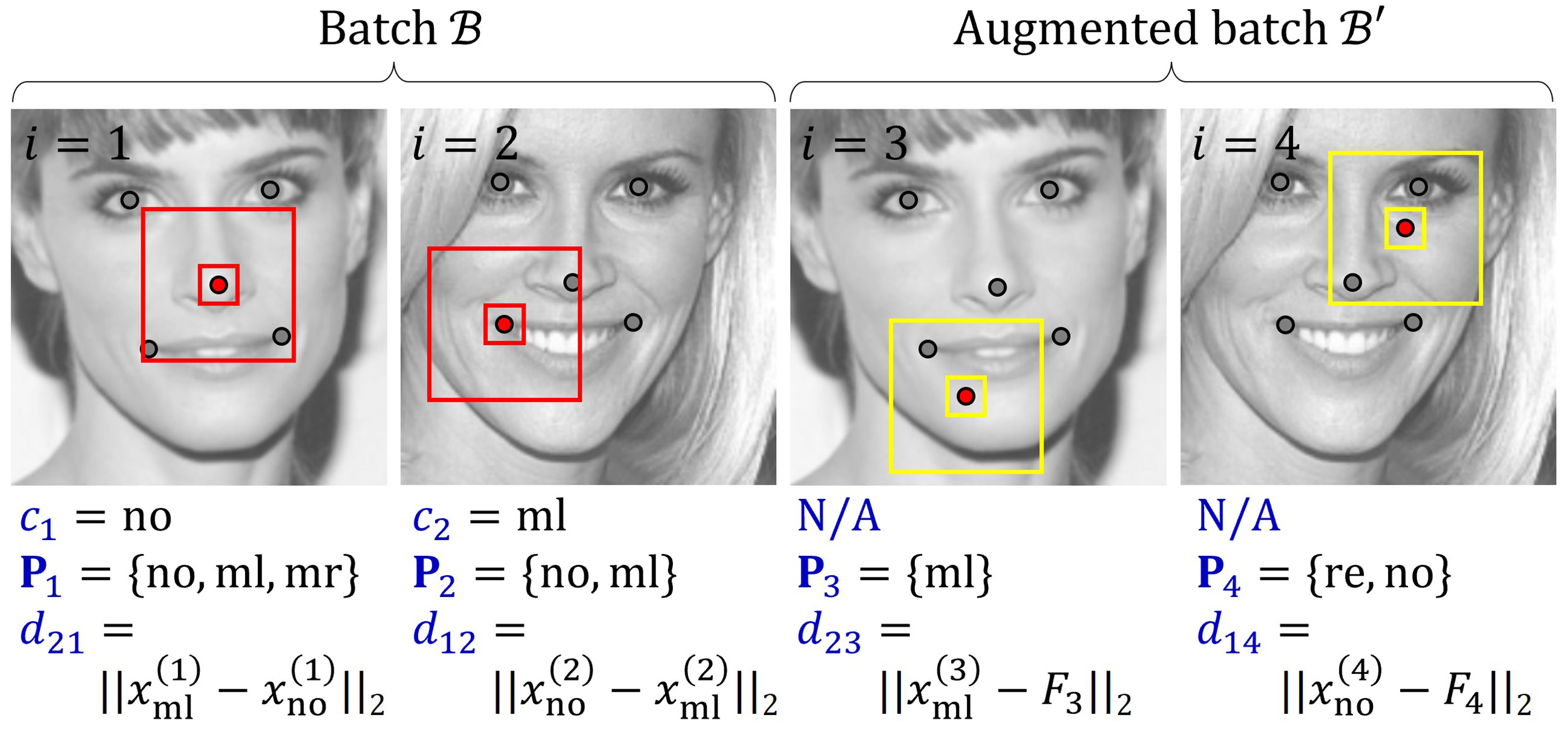}}
    \caption{\label{fig:augment}Example of sample augmentation for proximity-weighted contrastive learning. The right eye, left eye, nose, left mouth corner, and right mouth corner are denoted by \texttt{re}, \texttt{le}, \texttt{no}, \texttt{ml}, and \texttt{mr}, respectively.}
  \end{center}
  \vskip -0.1in
\end{figure}

\textbf{Local observation as input.}
An AI agent is assumed to observe the 2D world through two receptive fields, $\mathrm{RF}^{[1]}$ and $\mathrm{RF}^{[2]}$, of different sizes but centered at the same fixation point $\boldsymbol{F}\in\mathbb{R}^2$, as illustrated in Fig.~\ref{fig:augment}. This design ensures that each observation captures both fine-grained appearance cues and broader contextual information. 
Specifically, the observation $\boldsymbol{o}$ is constructed by extracting two patches from an input image of size $C\times H\times W$: 
\begin{equation*}
\begin{aligned}
&o^{[1]}\in\mathbb{R}^{C\times a\times a} \textrm{\quad from\quad} \mathrm{RF}^{[1]}\text{,}\\ &o^{[2]}\in\mathbb{R}^{C\times sa\times sa} \textrm{\quad from\quad} \mathrm{RF}^{[2]}\text{,}
\end{aligned}
\end{equation*}
where $s>1$ and $a\ll H, W$. The larger-scale patch $o^{[2]}$ is resized to $C\times a\times a$ and concatenated with $o^{[1]}$ along the channel dimension to form the final observation $\boldsymbol{o}\in\mathbb{R}^{2C\times a\times a}$. In this work, we set $a=27$ and $s=4$. When a receptive field extends beyond the image boundary, constant-value padding is applied to the out-of-bound regions.

\textbf{Spatio-semantic latent vector as output.}
PdEnc encodes observation $\boldsymbol{o}\in\mathbb{R}^{2C\times a\times a}$ for a given fixation point $F$ into a spatio-semantic latent vector $\boldsymbol{z}\in\mathbb{R}^{d_z}$, where $d_z=256$. This spatio-semantic latent vector lies on a hypersphere, $\boldsymbol{z}^\textrm{T}\boldsymbol{z}=1$.

\textbf{PdEnc architecture.}
The network architecture of PdEnc is illustrated in Fig.~\ref{fig:architectures}\textbf{a}. PdEnc is a simple convolutional neural network (CNN) with six convolutional layers and two fully connected layers: 2C32(6C32)-32C32-32C64-64C64-64C128-128C256-FC512-FC256-L2Norm for grayscale(RGB) images. The final L2Norm operation constrains latent representations to lie on a hypersphere, which stabilizes learning and facilitates distance-based reasoning in the spatio-semantic latent space.

\textbf{Multiscale receptive fields.}
The spatial proximity is defined as the real-world distance between the fixation point and surrounding objects within the second-scale receptive field $\mathrm{RF}^{[2]}$. This distance provides a continuous measure of spatial closeness, enabling PdEnc to learn latent representations that reflect both semantic similarity and fixation-centered spatial structure. It is important to note that spatial proximity can only be reliably inferred within the support of the receptive field since any distal objects outside the receptive field do not provide sufficient evidence for estimating spatial proximity.

\textbf{Proximity-weighted contrastive learning.}
To train PdEnc to explicitly enforce the semantic and spatial structures described above, we propose a proximity-weighted contrastive learning method. This method uses multiple image samples, each of which includes multiple objects $c$ with their coordinates $\boldsymbol{x}_{c}$ annotated.
\begin{equation*}\label{equ:c_def}
\boldsymbol{\mathrm{C}}_i=\left\{c|\textrm{objects $c$ included in image  $i$}\right\}\text{.}
\end{equation*}
For training, each mini-batch $\mathcal{B}$ is constructed by (i) randomly sampling images from a given dataset, (ii) randomly sampling a single object $c$ (along with their annotated coordinates $\boldsymbol{x}_c$) in $\boldsymbol{\mathrm{C}}_i$ for each image, and (iii) subsequently constructing observation $\boldsymbol{o}\in\mathbb{R}^{2C\times a\times a}$ by placing the fixation point $F$ on the object coordinate $\boldsymbol{x}_c$ $(F=\boldsymbol{x}_c)$ for each sampled object $c$. For the $i$th sample (object $c_i$ and $F_i=\boldsymbol{x}_{c(i)}$), we define a set of proximal objects:  
\begin{equation}\label{equ:prox_obj_set}
\boldsymbol{\mathrm{P}}_{i}=\left\{c\in\boldsymbol{\mathrm{C}}_i|\boldsymbol{x}_c\in o^{[2]}\text{ for }F_i\textrm{ on image }i\right\}\text{.}
\end{equation}
An augmented mini-batch $\mathcal{B}'$ is constructed by (i) randomly placing a single fixation point $F$ on each image in the batch $\mathcal{B}$ and (ii) subsequently constructing observation $\boldsymbol{o}$ for each image. A set of proximal objects for each sample in $\mathcal{B}'$ is also defined as in Eq.~\eqref{equ:prox_obj_set}. 

The goal of proximity-weighted contrastive learning is two-fold: 
\begin{itemize}
    \item Optimal mapping of objects with regard to their identities and spatial proximity using the observations (with $F=\boldsymbol{x}_c$) in batch $\mathcal{B}$
    \item Optimal mapping of random observations with respect to their proximal objects using the observations (with random $F$) in augmented batch $\mathcal{B}'$
\end{itemize}
To this end, we introduce proximity-weighted contrastive loss (PWConLoss) as follows.
\begin{equation}\label{equ:pwconloss}
\begin{aligned}
&\mathcal{L}_\text{PWC}=\frac{-1}{|\mathcal{B}|} \sum_{i\in\mathcal{B}}\biggr[\underbrace{\dfrac{1}{N_i}\sum_{j\in\mathcal{B}\backslash \left\{i\right\}}w_{ij}\mathds{1}_{\left\{c_i\in\boldsymbol{\mathrm{P}}_j\right\}}l_{ij}}_\text{between $c_i$ and $c_j$ ($i\neq j$)}\\
&\qquad\qquad\quad\qquad+\underbrace{\dfrac{1}{N'_i}\sum_{j\in\mathcal{B}'}w_{ij}\mathds{1}_{\left\{c_i\in\boldsymbol{\mathrm{P}}_j\right\}}l_{ij}}_\text{between $c_i$ and random observation $\boldsymbol{o}$}\biggr]\text{.}
\end{aligned}
\end{equation}
\begin{equation*}
\begin{aligned}
&N_i=\sum_{j\in\mathcal{B}\backslash \left\{i\right\}}\mathds{1}_{\left\{c_i\in\boldsymbol{\textrm{P}}_j\right\}}\text{,\quad}N'_i=\sum_{j\in\mathcal{B}'}\mathds{1}_{\left\{c_i\in\boldsymbol{\textrm{P}}_j\right\}}\text{,}\\
&l_{ij} = \log\frac{\exp(\boldsymbol{z}_i^\textrm{T}\boldsymbol{z}_j/\tau)}
{\sum\limits_{k\in(\mathcal{B}\cup\mathcal{B}')\backslash \left\{i\right\}}\exp(\boldsymbol{z}_i^\textrm{T}\boldsymbol{z}_k/\tau)}\text{,}
\end{aligned}
\end{equation*}
where $\boldsymbol{z}_i$ and $\tau$ denote the output vector for the $i$th sample and temperature parameter, respectively. 
The weight $w_{ij}$ is constrained to range $(1,2]$ such that 
\begin{equation}\label{equ:wgt_encoder}
w_{ij}=1+\exp\left(-0.025d_{ij}\right)\text{,}
\end{equation}
where $d_{ij}=||\boldsymbol{x}_{c(i)}^{(j)} - F_j||_2$. Note that $\boldsymbol{x}_{c(i)}^{(j)}$ means the coordinate of object $c_i$ on the $j$th sample.  
The constant $0.025$ is chosen so that $w_{ij}\approx1.5$ when $d_{ij}=27$, which corresponds to the height and width of the smaller receptive field $o^{[1]}$ with size $C\times27\times27$. Examples of sample augmentation for proximity-weighted contrastive learning are shown in Fig.~\ref{fig:augment}.

\subsection{Localizer}
\textbf{Localizer input.}
Localizer (Loc) infers the coordinates of proximal objects in $\boldsymbol{\textrm{P}}$ in Eq.~\eqref{equ:prox_obj_set} from the spatio-semantic output $\boldsymbol{z}$ of PdEnc for a given fixation point $F$. Assume that all objects in a given dataset ($\boldsymbol{\textrm{C}}^\textrm{tot}=\bigcup_{i=1}^{N_\textrm{s}}\boldsymbol{\textrm{C}}_i$; $N_\textrm{s}$ is the total number of samples) are included in a set of proximal objects for any fixation point. If this holds, a single observation is sufficient to localize all objects. However, since observation is made through \textit{local} receptive fields over an entire image, this assumption unlikely holds. Therefore, Loc collects $N_\text{o}$ observations (for different fixation points) that, in aggregate, cover all object classes in $\boldsymbol{\textrm{C}}^\textrm{tot}$. A pair of a fixation-point and the resulting latent representation $(\boldsymbol{z},F)$ is concatenated across total $N_\text{o}$ observations and subsequently vectorized to construct Loc input $\boldsymbol{I}\in \mathbb{R}^{N_\text{o}(d_z+2)}$.

\textbf{Localizer architecture.} The network architecture of Loc is illustrated in Fig.~\ref{fig:architectures}\textbf{b}.
Loc is designed to simultaneously predict the object coordinates normalized to the range $(-1,1)$ for all $|\boldsymbol{\textrm{C}}^\text{tot}|$ object classes, 
$\hat{\boldsymbol{x}}\in\mathbb{R}^{2|\boldsymbol{\textrm{C}}^\text{tot}|}$. It is a simple multilayer perceptron of FC512-FC512-FC$n$-Tanh, where $n=2|\boldsymbol{\textrm{C}}^\text{tot}|$.  
Rather than producing independent predictions per observation, Loc jointly aggregates all latent representations to infer object coordinates for each class. By leveraging the spatio-semantic structure encoded in $\boldsymbol{z}$, Loc enables localization of class-specific objects from a small set of local observations, without relying on pixel-wise computation or global image access.

\subsection{Auxiliary Localizer}
\textbf{Auxiliary localizer input.} Auxiliary localizer (AuxLoc) is optional model that refines the object coordinates $\hat{\boldsymbol{x}}$ inferred by Loc for each object separately. For the predicted coordinate $\hat{\boldsymbol{x}}$ in each object class, a first-scale image patch $o^{[1]}\in\mathbb{R}^{C\times a\times a}$ for $F=\hat{\boldsymbol{x}}$ is extracted from the input image. An object class-conditioned embedding is constructed as a tensor of size $1\times a \times a$, with values normalized to the range $[-1,1]$, and concatenated with $o^{[1]}$ along the channel dimension, yielding AuxLoc input $\boldsymbol{I}\in\mathbb{R}^{(C+1)\times a\times a}$. 

\textbf{Auxiliary localizer architecture.} The network architecture of AuxLoc is illustrated in Fig.~\ref{fig:architectures}\textbf{c}. AuxLoc is a CNN of 2C24(4C24)-4$\times$DW24-C1 for grayscale(RGB) images, where DW denotes a depth-wise separable convolution block~\cite{chollet2017depthwise}. In response to the local view centered at $\hat{\boldsymbol{x}}$, AuxLoc outputs a class-specific heatmap $\boldsymbol{h} \in \mathbb{R}^{1 \times a \times a}$. The refined object location is obtained by applying an offset derived from the peak of the heatmap $\boldsymbol{h}$ to the corresponding coarse estimate $\hat{\boldsymbol{x}}$. 

\section{Experimental Results}
As a proof-of-concept, we applied the WorldComp2D framework to facial landmark localization tasks in which facial landmarks are the objects whose spatio-semantic representations are learned. 
We used the COFW~\cite{burgos2013cofw}, 300W~\cite{sagonas2016300w}, and AFLW~\cite{koestinger2011aflw} datasets. The COFW dataset consists of 1,345 training and 507 test grayscale images, each annotated with 29 facial landmarks ($|\boldsymbol{\textrm{C}}^\text{tot}|=29$). COFW frequently contains facial landmarks occluded by external objects. The 300W dataset contains 3,148 training and 689 test RGB images, each annotated with 68 landmarks ($|\boldsymbol{\textrm{C}}^\text{tot}|=68$), and exhibits substantial variations in facial pose and illumination conditions. The AFLW dataset comprises 20,000 training and 4,386 test RGB images, each annotated with 19 landmarks ($|\boldsymbol{\textrm{C}}^\text{tot}|=19$), and is characterized by large pose variations and partial facial visibility.

\subsection{Implementation Detail}
Each image was cropped to include the full head, randomly rescaled ($\pm5\%$), horizontally flipped ($50\%$), and rotated ($60\%,\pm10\degree$), then resized to $256\times256$. During facial landmark localization, we first computed the mean location for each landmark across the samples in a given dataset, and Loc predicted an offset relative to this mean. To improve robustness, random offsets uniformly sampled from $[-5, 5]$ pixels were applied to the fixation points when training Loc and AuxLoc. The ground-truth heatmap for AuxLoc is constructed as a 2D Gaussian centered at the ground-truth landmark coordinate within the extracted patch $o^{[1]}$, with a fixed standard deviation of 1.5. The samples whose ground-truth coordinates fall outside the extracted patch $o^{[1]}$ were ignored during training. 

\begin{figure}[tb]
  \begin{center}
    \centerline{\includegraphics[width=\columnwidth]{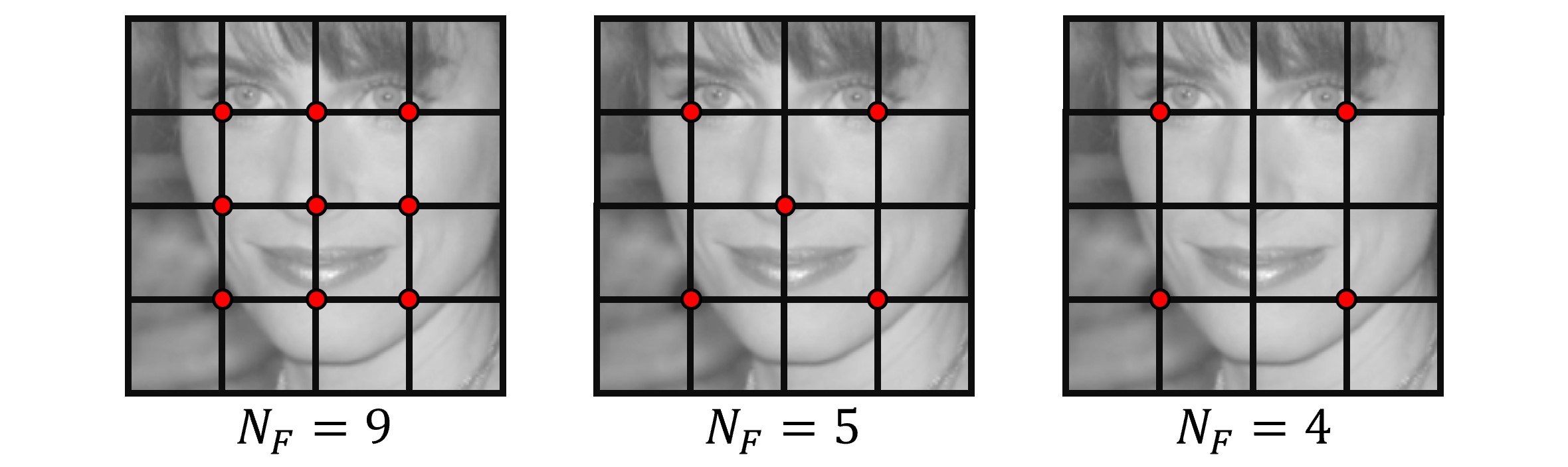}}
    \caption{\label{fig:fixation_points} Fixation points on a given image for $N_\text{F}=9$, $5$, and $4$.}
  \end{center}
  \vskip -0.1in
\end{figure}

We set the number of fixation points to nine ($N_\text{F}=9$) to cover all landmarks on each $256\times256$ sample, evenly distributed at fixed spatial intervals of 64 pixels along both image axes. Examples of $N_\text{F}=9/5/4$ are shown in Fig.~\ref{fig:fixation_points}.
The coordinate refinement by AuxLoc was constrained to at most 2 pixels on COFW and 1 pixel on other datasets.
The models were trained using the Pytorch framework~\citep{paszke2019pytorch} on a GPU workstation (RTX A6000; Xeon Gold CPU 2.9GHz; 256GB DRAM). The hyperparameters and learning behaviors of PdEnc, Loc, and AuxLoc are detailed in Appendix.

\subsection{Analysis of proximity-dependent encoder}

\begin{figure}[tb]
  \begin{center}
    \centerline{\includegraphics[width=\columnwidth]{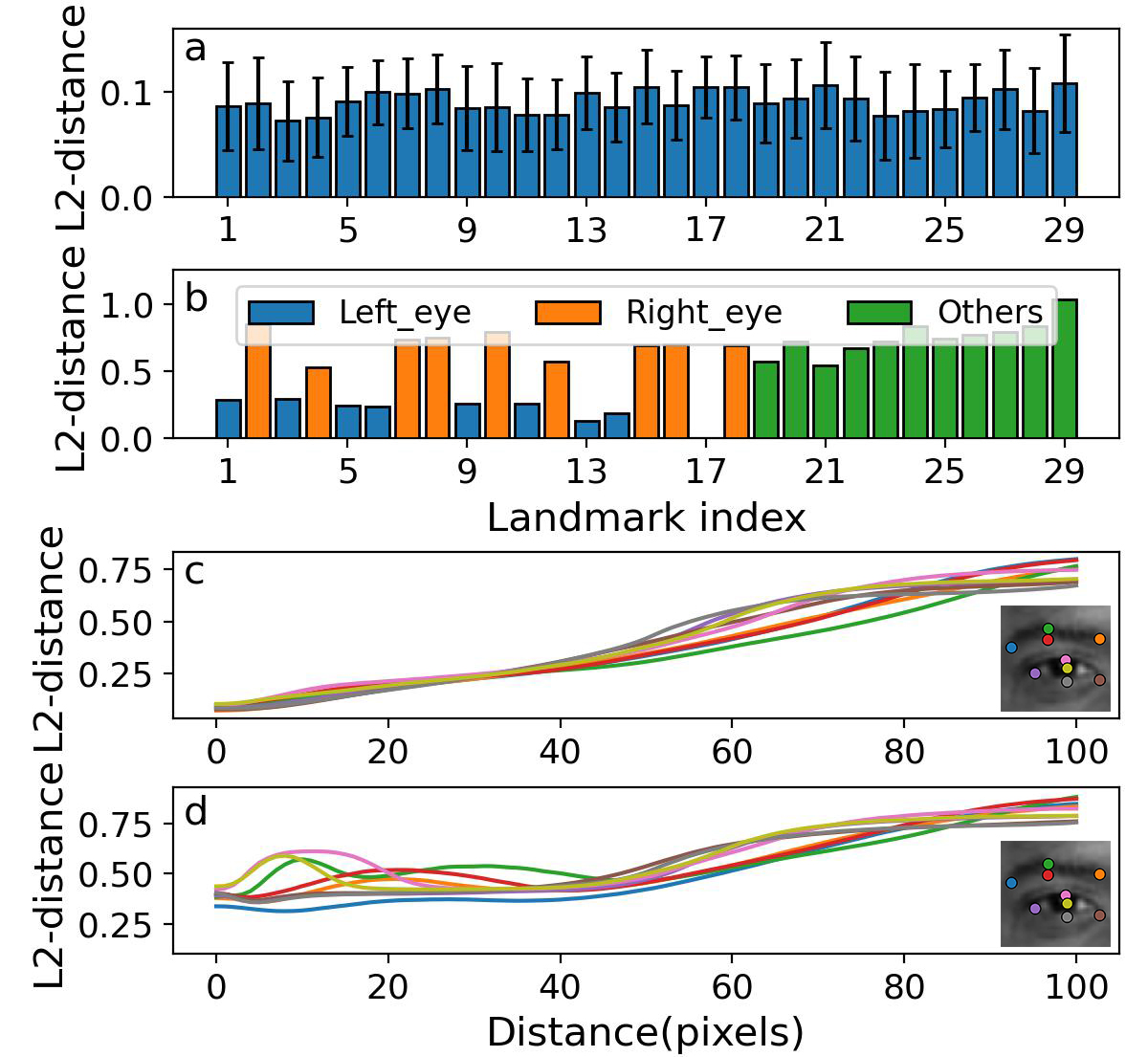}}
    \caption{\label{fig:l2_distance_encoder} Analysis of spatio-semantic latent representation. (\textbf{a}) L2 distances between individual landmark representations and their corresponding class means. (\textbf{b}) L2 distances between the left pupil representation and other landmarks representations within the same image. L2 distances as a function of spatial distance from a given landmark obtained from (\textbf{c}) PdEnc and (\textbf{d}) from an encoder trained using proximity-\textit{unweighted} contrastive loss.}
  \end{center}
  \vskip -0.1in
\end{figure}

\textbf{Intra-class clustering.} We evaluated the L2 distances between individual spatio-semantic representations for each class on COFW (with 29 landmarks). Their mean values are plotted in Fig.~\ref{fig:l2_distance_encoder}\textbf{a}, indicating successful class-wise clustering. This result suggests that the latent space effectively preserves landmark identity, yielding stable landmark-specific representations across different spatial contexts.

\textbf{Inter-class separation.} Fig.~\ref{fig:l2_distance_encoder}\textbf{b} shows the L2 distances between the left pupil (Class 17) representation and other landmark representations within the same image. While clear separation is observed across landmark clusters, landmarks that are spatially adjacent to the left pupil, such as other left eye related landmarks, exhibit relatively smaller distances. This pattern indicates that the latent space distinguishes landmark identity while simultaneously encoding relative spatial relationships among landmarks.

\textbf{Distance-preserving embedding.} We examined whether the real-world distance between an object and the fixation point is preserved in the spatio-semantic latent space (i.e., whether latent-space distance reflects physical proximity) by sampling observations on a fixation point with controlled offsets relative to the left eye landmark. 
As shown in Fig.~\ref{fig:l2_distance_encoder}\textbf{c}, L2 distances increase with real-world distance, indicating that spatial proximity is preserved in the latent space. Within the range of the second-scale receptive field (approximately 54 pixels), this pattern remains consistent across landmarks. Beyond this range, real-world proximity can no longer be reliably inferred, resulting in increased landmark-dependent variability in latent distances. 

As a counter part, we trained an encoder of the same architecture as PdEnc but using proximity-\textit{unweighted} contrastive loss that is the same as Eq.~\eqref{equ:pwconloss} but with $w_{ij}=1$ for all $i$ and $j$. Fig.~\ref{fig:l2_distance_encoder}\textbf{d} shows the latent space L2 distance between a given object and a fixation point with real-world distance. Compared with Fig.~\ref{fig:l2_distance_encoder}\textbf{c}, the spatial proximity is obviously unpreserved.

\subsection{Results on Facial Landmark Localization}

\begin{figure*}[ht]
  \begin{center}
    \centerline{\includegraphics[width=2\columnwidth]{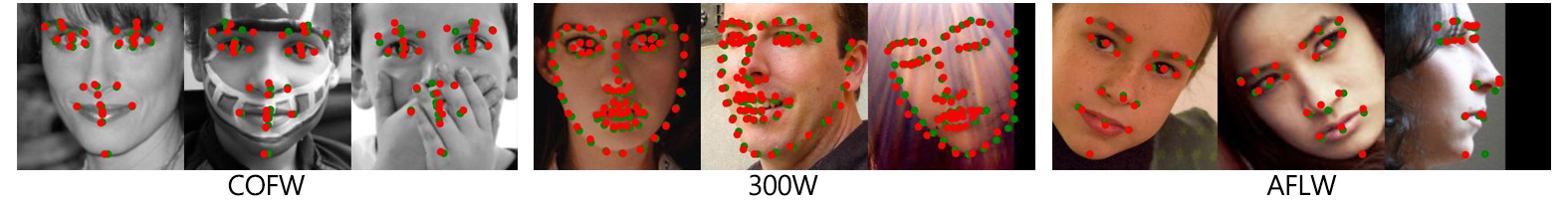}}
    \caption{\label{fig:visualization} Localized landmarks (red circles) and ground‑truth annotations (green circles) on sample images from COFW, 300W and AFLW.}
  \end{center}
  \vskip -0.1in
\end{figure*}

\begin{table*}[tb!]
\caption{Comparison of our method with SoTA approaches on COFW, 300W and AFLW.}
\label{tab:performance_comp}
\begin{center}
\setlength{\tabcolsep}{4.5pt}
\begin{small}
\begin{tabular}{lcccccc}
    \hline
    \multirow{2}*{Method} & \multicolumn{2}{c}{COFW} & 300W & AFLW & \multirow{2}*{$\#$ Params (M)} & \multirow{2}*{FLOPs} \\
    \cline{2-5}
    & NME$_\mathrm{IO}$ & NME$_\mathrm{IP}$ & NME$_\mathrm{IO}$ & NME$_\mathrm{Diag}$\\
    \hline
    LAB~\cite{wu2018lab} & 3.92 & 5.58 & 3.49 & 1.85 & 25.1 & 18.9G \\ 
    AWing~\cite{wang2019awing} & - & 4.94 & 3.07 & 1.53 & 24.2 & 26.8G \\
    AVS~\cite{qian2019avs} & - & 4.43 & 3.86 & 1.86 & 28.3 & 2.4G \\
    ODN~\cite{zhu2019odn} & - & - & 4.17 & 1.63 & - & - \\
    HRNet~\cite{wang2020hrnet} & 3.45 & - & 3.32 & 1.57 & 9.7 & 4.8G \\
    PIP~\cite{jin2021pip} & 3.45 & - & 3.19 & 1.42 & 45.7 & 10.5G \\
    ADNet~\cite{huang2021adnet} & - & 4.69 & 2.93 & - & 13.4 & 17.0G \\
    SDFL~\cite{lin2021sdfl} & 3.63 & - & - & - & - & 5.2G \\
    HIH~\cite{lan2021hih} & 3.21 & 4.63 & 3.09 & - & 22.7 & 17.2G \\
    SLPT~\cite{xia2022slpt} & 3.32 & 4.63 & 3.17 & - & 13.2 & 6.1G \\
    STARLoss~\cite{zhou2023star} & - & 4.62 & 2.87 & - & 13.4 & - \\
    RHT-R~\cite{wan2023rht} & - & 4.42 & 2.82 & 1.18 & - & - \\
    D-ViT~\cite{dang2025dvit} & - & 4.13 & 2.85 & - & 67.3 & 21.8G \\
    PoPos~\cite{xiang2025popos} & - & 3.80 & 3.28 & 1.43 & 9.7 & 1.2G \\
    \hline
    \makecell[l]{\textbf{WorldComp2D on} \\ \quad \textbf{COFW/300W/AFLW}} & 5.16$\pm$0.05 & 7.43$\pm$0.07 & 5.06$\pm$0.01 & 1.52$\pm$0.01 & 2.4 & \makecell{293.7M / 546.8M / 256.9M} \\
    \hline
\end{tabular}
\end{small}
\end{center}
\vskip -0.1in
\end{table*}

\begin{table*}[tb!]
\caption{FPS and tolerance to reduced data precision.}
\label{tab:precision}
\begin{center}
\begin{small}
\begin{tabular}{lcccccc}
    \hline
    \multirow{2}*{Precision} & \multicolumn{2}{c}{COFW} & \multicolumn{2}{c}{300W} & \multicolumn{2}{c}{AFLW} \\
    \cline{2-7}
     & FPS & NME$_\mathrm{IO}$ & FPS & NME$_\mathrm{IO}$ & FPS & NME$_\mathrm{Diag}$ \\ \hline
     FP32 on CPU i5 & 138.4 & 5.16$\pm$0.05 & 78.48 & 5.06$\pm$0.01 & 163.55 & 1.52$\pm$0.01  \\
     \hline
    FP32 on GPU A6000 & 9328.96 & 5.16$\pm$0.05 & 5905.59 & 5.06$\pm$0.01 & 6381.85 & 1.52$\pm$0.01  \\
    FP16 on GPU A6000 & 9556.62 & 5.16$\pm$0.05 & 7217.09 & 5.06$\pm$0.02 & 7709.42 & 1.52$\pm$0.01  \\
    BF16 on GPU A6000 & 9834.96 & 5.17$\pm$0.04 & 7486.92 & 5.07$\pm$0.02 & 7743.49 & 1.52$\pm$0.01\\    
    \hline 
\end{tabular}
\end{small}
\end{center}
\vskip -0.1in
\end{table*}

\textbf{Localization accuracy and efficiency.} Fig.~\ref{fig:visualization} presents landmark localization results on several samples from COFW, 300W, and AFLW. 
The localization accuracy is measured using the Normalized Mean Error (NME), where the error is normalized by the inter-ocular distance (NME$_\mathrm{IO}$) or inter-pupil distance (NME$_\mathrm{IP}$) or the diagonal length of the face bounding box (NME$_\mathrm{Diag}$), following standard evaluation protocols for each dataset. Since WorldComp2D is not trained via direct supervised regression for landmark localization, it yields higher NME values on COFW and 300W compared to regression-based SoTA methods listed in Table~\ref{tab:performance_comp}. In contrast, on the AFLW, our framework achieves lower NME than LAB~\cite{wu2018lab}, Awing~\cite{wang2019awing}, ODN~\cite{zhu2019odn}, and HRNet~\cite{wang2020hrnet}. 

Beyond localization accuracy, WorldComp2D highlights extremely low computational complexity. The entire framework consists of 2.4M parameters in total, with 1.1M, 1.3M, and 4.0K parameters assigned to PdEnc, Loc, and AuxLoc, respectively. In terms of computational cost, PdEnc, Loc, and AuxLoc require approximately 15.7M, 3.0M, and 5.9M FLOPs, respectively. Since the framework processes nine fixation-centered observations ($N_\text{F}=9$), the total number of FLOPs is computed as 
\begin{equation*}
 \text{FLOPs$_\text{tot}$} = 9\text{FLOPs$_\text{PdEnc}$} + \text{FLOPs$_\text{Loc}$} + |\boldsymbol{\textrm{C}}^\text{tot}|\text{FLOPs$_\text{AuxLoc}$}\text{,} 
\end{equation*}
resulting in 293.7, 546.8, and 256.9 MFLOPs on COFW, 300W, and AFLW, respectively.
Compared to the lightweight PoPos~\cite{xiang2025popos} model with 9.7M parameters, WorldComp2D achieves a 4.0$\times$ reduction in the number of parameters and a 2.2$\times$ reduction in FLOPs in the worst case. 

\textbf{Real-time operation.} 
Due to its low computational complexity, our method runs at 138.41 (COFW), 78.48 (300W), and 163.55 (AFLW) frames per second (FPS) on a desktop (i5-13400 CPU 2.5GHz; 128GB DRAM), using a batch size of 1 (Table~\ref{tab:precision}). This highlights the feasible real-time operation of WorldComp2D.

\begin{figure}[t]
  \begin{center}
    \centerline{\includegraphics[width=\columnwidth]{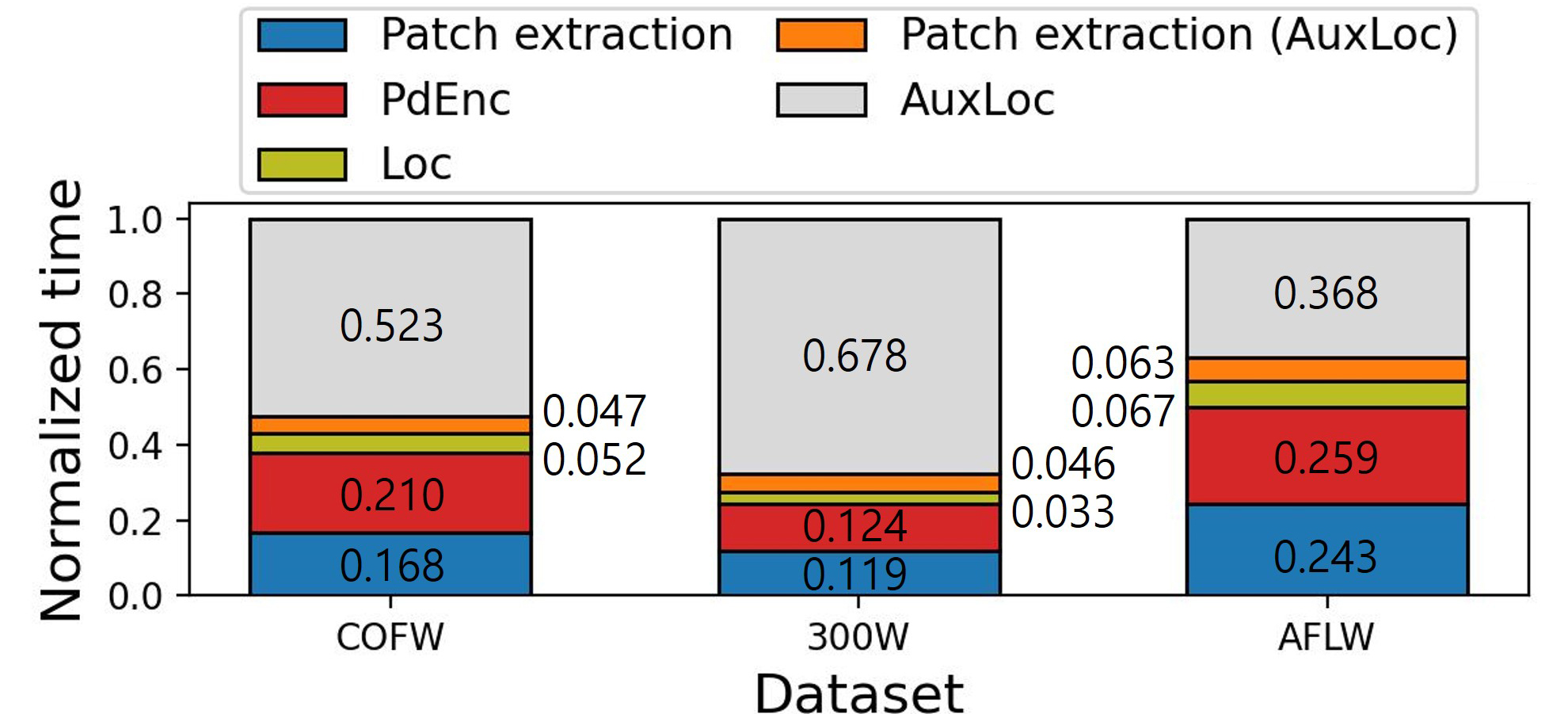}}
    \caption{\label{fig:stage_time} Normalized localization runtime decomposed into five sub-workloads.}
  \end{center}
  \vskip -0.1in
\end{figure}

We decompose the overall framework into five sub-workloads: (1) patch extraction (local observation for a given fixation point for PdEnc), (2) PdEnc operation, (3) Loc operation, (4) optional patch extraction for AuxLoc, and (5) AuxLoc operation. 
Fig.~\ref{fig:stage_time} shows the normalized runtime for each of the five workloads. Across all datasets, AuxLoc occupies the largest contribution to the total runtime, followed by PdEnc and patch extraction stages. This is because AuxLoc (despite lightweight) processes a separate local patch for each landmark. Accordingly, its contribution decreases as the number of landmarks decreases. 
PdEnc accounts for the larger contribution than Loc as it operates on multiple fixation-centered patches per frame, whereas Loc relies on a single forward pass on aggregated representations. 
Patch extraction constitutes another non-negligible component since it involves patching and resizing for the second-scale receptive field ($o^{[2]}$). This leads to a longer processing time than patch extraction for AuxLoc that uses first-scale patches only.

\begin{table}[tb!]
\caption{Effect of AuxLoc on localization accuracy and computational efficiency on CPU i5.}
\label{tab:ablation_auxloc}
\begin{center}
\begin{small}
\begin{tabular}{lccc}
    \hline
    \multicolumn{4}{c}{COFW} \\ \hline
    Model & NME$_\mathrm{IO(Diag)}$ & FLOPs & FPS \\ \hline
    with AuxLoc & 5.16$\pm$0.05 & 293.7M & 138.41 \\
    w/o AuxLoc & 5.41$\pm$0.04 & 140.5M & 344.19 \\ \hline
    \multicolumn{4}{c}{300W} \\ \hline
    with AuxLoc & 5.06$\pm$0.01 & 546.8M & 78.48 \\
    w/o AuxLoc & 5.22$\pm$0.01 & 144.7M & 282.34 \\ \hline
    \multicolumn{4}{c}{AFLW} \\ \hline
    with AuxLoc & 1.52$\pm$0.01 & 256.9M & 163.55 \\
    w/o AuxLoc & 1.61$\pm$0.01 & 144.6M & 304.93 \\ \hline
\end{tabular}
\end{small}
\end{center}
\vskip -0.1in
\end{table}

As such, AuxLoc serves as an auxiliary module that refines the localization predicted by Loc, but it still accounts for the largest contribution to the overall runtime. Table~\ref{tab:ablation_auxloc} reports the accuracy improvement from adding AuxLoc and its computational overhead. In the best case, AuxLoc yields only a 4.6\% reduction in NME on COFW, underscoring the strong localization accuracy of the baseline (Loc-only) model. However, omitting AuxLoc during object localization substantially reduces FLOPs and correspondingly improves FPS. This FPS gain becomes more pronounced as the number of landmarks increases, since AuxLoc causes additional per-landmark computation.   

\textbf{Effect of data precision.} 
We examined the localization accuracy and computational efficiency for three different data formats (FP32, FP16, and BF16) on GPU. As summarized in Table~\ref{tab:precision}, FP16 and BF16 substantially improve FPS while leading to only negligible degradation in NME. These results highlight that WorldComp2D can be deployed efficiently under practical deployment constraints.

\begin{table}[tb!]
\caption{Localization accuracy under various degradations.}
\label{tab:robustness}
\begin{center}
\begin{small}
\setlength{\tabcolsep}{4.5pt}
\begin{tabular}{lccc}
    \hline
    Degradation & COFW & 300W & AFLW \\ \hline
    Baseline & 5.16$\pm$0.05 & 5.06$\pm$0.01 & 1.52$\pm$0.01  \\
    Blur ($\sigma=1$) & 5.15$\pm$0.05 & 5.06$\pm$0.02 & 1.53$\pm$0.01 \\
    Blur ($\sigma=2$) & 5.27$\pm$0.04 & 5.17$\pm$0.04 & 1.57$\pm$0.01  \\
    Blur ($\sigma=3$) & 5.60$\pm$0.02 & 5.43$\pm$0.06 & 1.64$\pm$0.02  \\
    JPEG ($Q=80$) & 5.17$\pm$0.04 & 5.08$\pm$0.01 & 1.52$\pm$0.01  \\
    JPEG ($Q=60$) & 5.18$\pm$0.03 & 5.10$\pm$0.01 & 1.52$\pm$0.01  \\
    JPEG ($Q=40$) & 5.19$\pm$0.03 & 5.11$\pm$0.01 & 1.53$\pm$0.01  \\
    JPEG ($Q=20$) & 5.22$\pm$0.06 & 5.19$\pm$0.02 & 1.54$\pm$0.01   \\
    Motion Blur ($k=5$) & 5.17$\pm$0.06 & 5.09$\pm$0.03 & 1.53$\pm$0.01   \\
    Motion Blur ($k=10$) & 5.57$\pm$0.03 & 5.42$\pm$0.04 & 1.68$\pm$0.01   \\
    Occlusion ($\text{size}=20$) & 5.33$\pm$0.05 & 5.27$\pm$0.01 & 1.59$\pm$0.01 \\
    Occlusion ($\text{size}=40$) & 5.56$\pm$0.07 & 5.58$\pm$0.01 & 1.64$\pm$0.01  \\
    \hline 
\end{tabular}
\end{small}
\end{center}
\vskip -0.1in
\end{table}

\textbf{Robustness to input degradation.}
Since PdEnc encodes real-world objects, robustness to input degradations is a prerequisite for reliable downstream localization. To identify this robustness, we evaluated the localization performance of WorldComp2D under various degradation cases, including Gaussian blur, JPEG degradation, motion blur, and occlusion. 
For the occlusion experiments, a square patch (height$=$width$=$size) with zero-valued pixels was inserted at a random location on each test sample.
Table~\ref{tab:robustness} summarizes the results. Across all datasets and degradation types, WorldComp2D exhibits consistently moderate performance degradation. Notably, the largest absolute degradation is observed under extreme occlusion (size=40) on 300W, where NME increases by 10.2\% only. These results suggest that our spatio-semantic representations provide a stable foundation for robust inference under diverse real-world image degradations.

\subsection{Ablation study}

\textbf{PdEnc complexity versus accuracy.} 
\begin{table}[tb!]
\caption{PdEnc with different model complexities.}
\label{tab:encoder}
\begin{center}
\begin{small}
\begin{tabular}{lccc}
    \hline
    \multicolumn{4}{c}{COFW} \\ \hline
    Model & $\#$ Params & NME$_\mathrm{IO(Diag)}$ & FLOPs \\ \hline
    PdEnc (C9) & 1.4M & 5.80$\pm$0.05 & 162.3M \\
    PdEnc (C16) & 1.8M & 5.30$\pm$0.06 & 195.2M \\
    PdEnc (C32) & 2.4M & 5.16$\pm$0.05 & 293.7M \\ \hline
    \multicolumn{4}{c}{300W} \\ \hline
    PdEnc (C9) & 1.4M & 5.88$\pm$0.07 & 412.5M  \\
    PdEnc (C16) & 1.8M & 5.28$\pm$0.03 & 446.2M \\
    PdEnc (C32) & 2.4M & 5.06$\pm$0.01 & 546.8M \\ \hline
    \multicolumn{4}{c}{AFLW} \\ \hline
    PdEnc (C9) & 1.4M & 1.74$\pm$0.02  & 122.6M \\
    PdEnc (C16) & 1.8M & 1.57$\pm$0.01 & 156.4M \\
    PdEnc (C32) & 2.4M & 1.52$\pm$0.01 & 256.9M \\ \hline
    \hline
\end{tabular}
\end{small}
\end{center}
\vskip -0.1in
\end{table}
We investigated a relationship between PdEnc complexity (model size) and the downstream localization accuracy. To this end, we considered two additional lighter models (C9 and C16) to our PdEnc (referred to as C32). The architectural details for these models are addressed in Appendix. The localization accuracy and the number of total FLOPs with these models are compared with C32 in Table~\ref{tab:encoder}. The downstream localization accuracy tends to increase with the model complexity at the cost of a increase in computation number. Regarding this tradeoff, we chose C32 as our standard PdEnc that realizes competitive localization accuracy while remaining lightweight.

\textbf{Cross-dataset evaluation.} 
\begin{table}[tb!]
\caption{Cross-dataset evaluation result.}
\label{tab:cofw68}
\begin{center}
\setlength{\tabcolsep}{3pt}
\begin{small}
\begin{tabular}{lccc}
    \hline
    Model & 300W & COFW-68 & Degradation \\ \hline
    LAB~\cite{wu2018lab} & 3.49 & 4.62 & 32.4\% \\
    ODN~\cite{zhu2019odn} & 4.17 & 5.30 & 27.1\% \\ 
    PIP~\cite{jin2021pip} & 3.23 & 4.23 & 31.0\% \\ 
    WorldComp2D & 5.06$\pm$0.01 & 6.08$\pm$0.14 & 20.2\% \\
    \hline
\end{tabular}
\end{small}
\end{center}
\vskip -0.1in
\end{table}
The COFW, 300W, and AFLW datasets contain different numbers of annotated landmarks. For cross-dataset evaluation, we conducted zero-shot landmark localization on COFW-68~\cite{ghiasi2014cofw68} (COFW with 68 landmarks comparable to 300W with $|\boldsymbol{\textrm{C}}^\text{tot}|=68$) using the WorldComp2D framework trained on 300W. Table~\ref{tab:cofw68} reports the cross-validation result and several SoTA results. These regression-based SoTA methods suffer from a performance degradation of $27–32\%$, whereas WorldComp2D shows only a $20.2\%$ increase in NME.

\begin{table}[tb!]
\caption{Localization accuracy and efficiency with different numbers of fixation points.}
\label{tab:fixation}
\begin{center}
\begin{small}
\begin{tabular}{lccc}
    \hline
    \multicolumn{4}{c}{COFW} \\ \hline
    Model & $\#$ Params & NME$_\mathrm{IO(Diag)}$ & FLOPs \\ \hline
    $N_\text{F}=4$ & 1.6M & 5.75$\pm$0.06 & 216.0M \\
    $N_\text{F}=5$ & 1.8M & 5.30$\pm$0.06 & 231.6M \\
    $N_\text{F}=9$ & 2.4M & 5.16$\pm$0.05 & 293.7M \\ \hline
    \multicolumn{4}{c}{300W} \\ \hline
    $N_\text{F}=4$ & 1.7M & 5.60$\pm$0.04 & 466.8M  \\
    $N_\text{F}=5$ & 1.9M & 5.34$\pm$0.04 & 482.8M \\
    $N_\text{F}=9$ & 2.4M & 5.06$\pm$0.01 & 546.8M \\ \hline
    \multicolumn{4}{c}{AFLW} \\ \hline
    $N_\text{F}=4$ & 1.7M & 1.69$\pm$0.01 & 176.9M \\
    $N_\text{F}=5$ & 1.9M & 1.62$\pm$0.01 & 192.9M \\
    $N_\text{F}=9$ & 2.4M & 1.52$\pm$0.01 & 256.9M \\ 
    \hline
\end{tabular}
\end{small}
\end{center}
\vskip -0.1in
\end{table}

\textbf{Number of fixation points.} 
Loc uses the spatio-semantic representations for multiple fixation points ($N_\text{F}=9$) in aggregate as its input because a single \textit{local} observation cannot include all objects (landmarks) in a given dataset. As such, we applied $3\times3$ periodic fixation points on each input image in an attempt to cover all landmarks for each set of nine observations. We also evaluated Loc with $N_\text{F}=4$ and $5$ as illustrated in Fig.~\ref{fig:fixation_points}. 
Table~\ref{tab:fixation} summarizes the effect of $N_\text{F}$ on the localization accuracy on the three datasets and the number of total FLOPs. An increase in $N_\text{F}$ increases evidence for real-world objects to be mapped into the latent space, leading to a higher localization accuracy. Nevertheless, WorldComp2D achieves competitive localization accuracy even with $N_\text{F}=4$, while significantly reducing its computational complexity.

\section{Conclusion}
We propose WorldComp2D, a lightweight representation learning framework for embodied AI agents that must reason from local observations under limited compute. It encodes local views into a spatio-semantic latent space that jointly captures object identity and spatial proximity, enabling localization without exhaustive full-image processing and providing an efficient alternative to global-view architectures.

Using facial landmark localization as a proof-of-concept, we show that this latent space has meaningful geometry (intra-class clustering, inter-class separation, and distances consistent with real-world proximity). WorldComp2D achieves competitive accuracy and high inference speed with substantially reduced compute and lightweight models, while enabling accuracy–efficiency trade-offs via an optional refinement module (AuxLoc). More broadly, explicitly structured latent geometry offers a compact, interpretable foundation for scalable perception and spatial inference under resource constraints.

\section{Limitations and future work}
WorldComp2D is validated mainly on a controlled 2D landmark localization setting, so generalization to more diverse objects, clutter, occlusion, and larger viewpoint/scale changes remain uncertain. The method also assumes a fixed observation policy (periodic fixation points). Future work includes extending to 3D embodied settings, learning an adaptive fixation policy, and evaluating on broader benchmarks and edge-deployment targets.

\section*{Acknowledgments}
This research was supported by Institute of Information \& communications Technology Planning \& Evaluation (IITP) grants funded by the Koreagovernment (MSIT) (RS-2023-00229689 and RS-2023-00253914).

\section*{Impact Statement}
This paper presents work whose goal is to advance the field of machine learning. There are many potential societal consequences of our work, none of which we feel must be specifically highlighted here.

\bibliography{main}
\bibliographystyle{icml2026}

\newpage
\appendix
\onecolumn
\setcounter{figure}{0}
\setcounter{table}{0}

\section{Training PdEnc, Loc, and AuxLoc}
\begin{table}[hbt]
\caption{Hyperparameters used.}
\label{tab:training_setting}
\begin{center}
\begin{small}
\begin{tabular}{lccc}
    \hline
    & PdEnc & Loc & AuxLoc \\
    \hline
    Optimizer & Adam & Adam & Adam \\ \hline
    $\#$ Epochs & 2000 (600 on AFLW) & 1000 (400 on AFLW) & 1000 (400 on AFLW)  \\ \hline
    Batch size & 128 & 50 & 50 \\ \hline
    Initial lr & 1E-2 & 5E-4 & 5E-4 \\ \hline
    Lr decay & 0.1 & 0.1 & 0.1 \\ \hline
    Decay epoch & 1000 (300 on AFLW) & 500 (200 on AFLW) & 500 (200 on AFLW)\\ \hline
    \hline
\end{tabular}
\end{small}
\end{center}
\end{table}
\begin{figure}[htb]
  \vskip 0.2in
  \begin{center}
    \centerline{\includegraphics[width=.6\columnwidth]{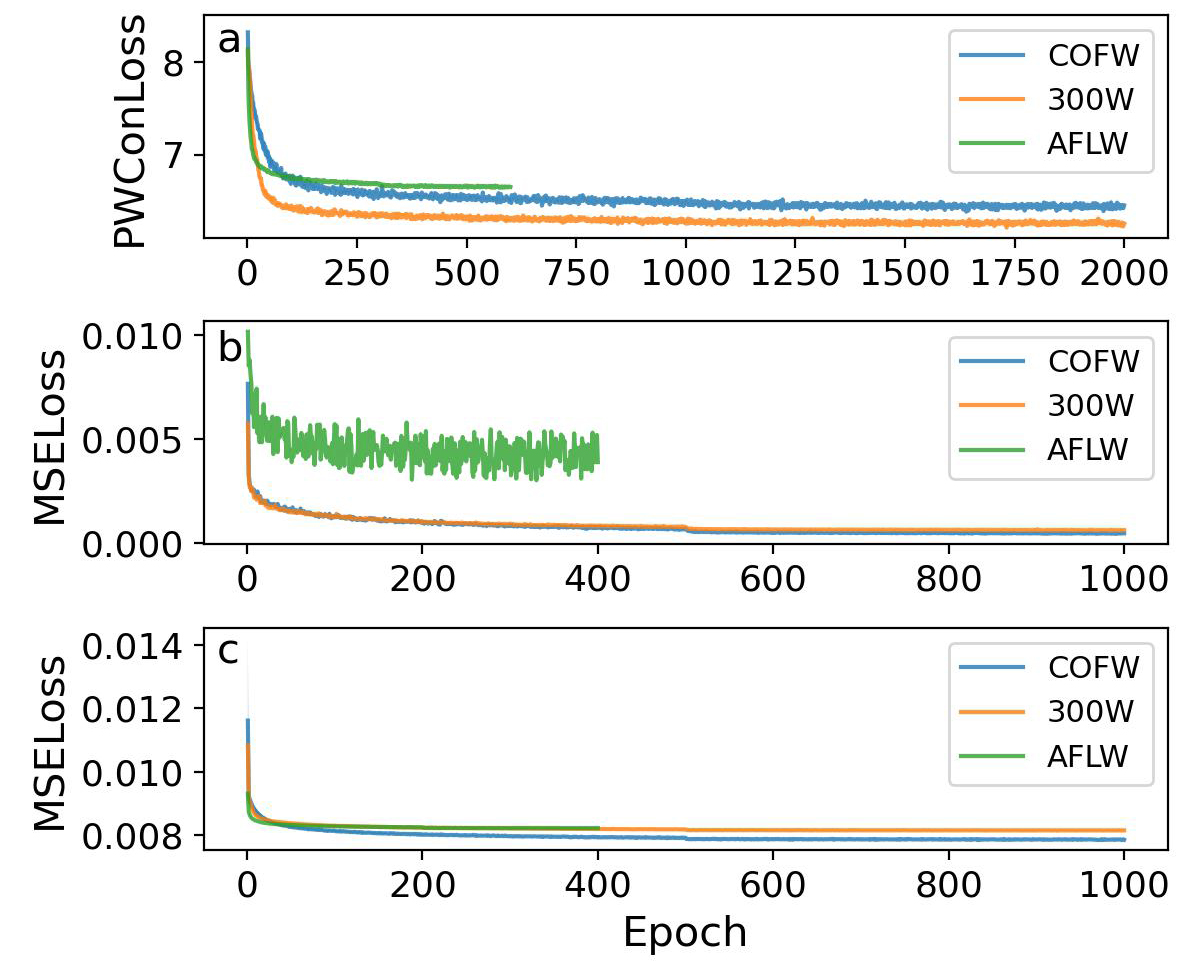}}
    \caption{\label{fig:learning_curve} Learning curve for \textbf{(a)} PdEnc, \textbf{(b)} Loc, and \textbf{(c)} AuxLoc.}
  \end{center}
\end{figure}
Table~\ref{tab:training_setting} lists the hyperparameters used to train PdEnc, Loc, and AuxLoc on the COFW, 300W, and AFLW datasets. The learning behavior of these elementary modules is plotted in Fig.~\ref{fig:learning_curve}.

\section{Summary of architectures for elementary modules}
\begin{table}[hbt]
\caption{Network architectures.}
\label{tab:architecture}
\begin{center}
\begin{small}
\begin{tabular}{c}
    \hline
    \textbf{PdEnc (C9)}\\ \hline
    2C9-9C16-16C32-32C64-FC256-FC128-L2Norm for COFW\\
    6C9-9C16-16C32-32C64-FC256-FC128-L2Norm for 300W and AFLW\\
    \hline
    \textbf{PdEnc (C16)}\\ \hline
    2C16-16C16($s=1$)-16C32-32C32($s=1$)-32C64-64C128-FC512-FC256-L2Norm for COFW\\ 
    6C16-16C16($s=1$)-16C32-32C32($s=1$)-32C64-64C128-FC512-FC256-L2Norm for 300W and AFLW \\ \hline
    \textbf{PdEnc (C32)}\\ \hline
    2C32-32C32($s=1$)-32C64-64C64($s=1$)-64C128-128C256-FC512-FC256-L2Norm for COFW \\
    6C32-32C32($s=1$)-32C64-64C64($s=1$)-64C128-128C256-FC512-FC256-L2Norm for 300W and AFLW \\\hline
    \textbf{Loc}\\ \hline
    FC512-FC512-FC$n$-Tanh ($n=58$ for COFW, $n=136$ for 300W, $n=38$ for AFLW)\\ \hline
    \textbf{AuxLoc}\\ \hline
    2C24-4*DW24-C1 for COFW \\
    4C24-4*DW24-C1 for 300W and AFLW \\\hline
    \hline
    \footnotesize{\makecell[l]{Unless otherwise specified, all convolution layers in PdEnc have the same convolutional operational settings, $k_H=k_W=3$, $s=2$, $p=1$.}}\\
\end{tabular}
\end{small}
\end{center}
\end{table}
Table~\ref{tab:architecture} lists the architectures of elementary modules on the COFW, 300W, and AFLW datasets.


\end{document}